\documentclass[letterpaper]{article} % DO NOT CHANGE THIS
\usepackage[submission]{aaai2026}  % DO NOT CHANGE THIS
\usepackage{times}  % DO NOT CHANGE THIS
\usepackage{helvet}  % DO NOT CHANGE THIS
\usepackage{courier}  % DO NOT CHANGE THIS
\usepackage[hyphens]{url}  % DO NOT CHANGE THIS
\usepackage{graphicx} % DO NOT CHANGE THIS
\usepackage{natbib}  % DO NOT CHANGE THIS AND DO NOT ADD ANY OPTIONS TO IT
\usepackage{caption} % DO NOT CHANGE THIS AND DO NOT ADD ANY OPTIONS TO IT
\usepackage{algorithm}
\usepackage{algorithmic}
\usepackage{pifont}

\usepackage{newfloat}
\usepackage{listings}
\DeclareCaptionStyle{ruled}{labelfont=normalfont,labelsep=colon,strut=off} % DO NOT CHANGE THIS
\floatstyle{ruled}
\newfloat{listing}{tb}{lst}{}
\floatname{listing}{Listing}
\title{\textit{GridCodex}: A RAG-Driven AI Framework for \\Power Grid Code Reasoning and Compliance}
\author{
    Jinquan Shi\textsuperscript{\rm 1},
    Yingying Cheng\textsuperscript{\rm 1},
    Fan Zhang\textsuperscript{\rm 1},
    Miao Jiang\textsuperscript{\rm 2},
    Jun Lin\textsuperscript{\rm 2},
    Yanbai Shen\textsuperscript{\rm 2}
}
\affiliations{
    \textsuperscript{\rm 1}Theory Lab, 2012 Labs, Huawei Technologies Co., Ltd.\\

    \textsuperscript{\rm 2}Digital Energy BU, Huawei Technologies Co., Ltd.\\

    \{shi.jinquan, cheng.yingying1, zhang.fan2, jiangmiao8, linjun37, shenyanbai\}@huawei.com
}

\usepackage{bibentry}
\begin{document}

\maketitle
\pubnote{Accepted at the 4th Deployable AI Workshop (DAI 2026), co-located with AAAI-26}

\begin{abstract}
    The global shift towards renewable energy presents unprecedented challenges for the electricity industry, making regulatory reasoning and compliance increasingly vital. Grid codes—the regulations governing grid operations—are complex and often lack automated interpretation solutions, which hinders industry expansion and undermines profitability for electricity companies. We introduce GridCodex, an end-to-end framework for grid code reasoning and compliance that leverages large language models and retrieval-augmented generation (RAG). Our framework advances conventional RAG workflows through multi-stage query refinement and enhanced retrieval with RAPTOR. We validate the effectiveness of GridCodex with comprehensive benchmarks, including automated answer assessment across multiple dimensions and regulatory agencies. Experimental results showcase a 26.4\% improvement in answer quality and more than a 10-fold increase in Recall@30. An ablation study further examines the impact of base model selection.
\end{abstract}

% Uncomment the following to link to your code, datasets, an extended version or similar.
% You must keep this block between (not within) the abstract and the main body of the paper.
% \begin{links}
%     \link{Code}{https://aaai.org/example/code}
%     \link{Datasets}{https://aaai.org/example/datasets}
%     \link{Extended version}{https://aaai.org/example/extended-version}
% \end{links}
\section{Introduction}

In the accelerating transition toward renewable energy, electricity has become the central medium linking diverse sustainable energy sources with end users, placing unprecedented demands on the robustness and efficiency of power grids \cite{GridCodeReinforcements2016,RenewablesIntegrationIslands2017,ntomarisReserveQuantificationInsular2014}. Grid codes—the rules and regulations that govern the reliable operation of power systems—specify critical technical requirements for integrating renewable energy resources, thereby safeguarding grid stability and security. Yet, these codes differ significantly across countries and regions \cite{ullahDynamicPerformancePower2025}, rendering grid code compliance a complex but indispensable challenge for electricity infrastructure providers seeking international expansion.

\begin{figure}[htbp!]
\centering
\includegraphics[width=0.8\columnwidth, trim=5cm 7cm 13cm 2cm, clip]{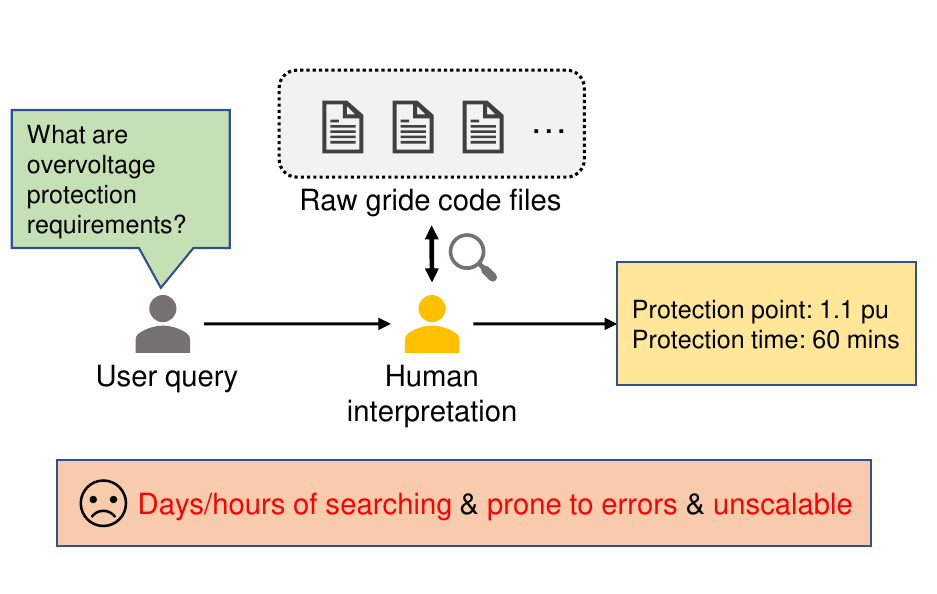}
\caption{Conventional human grid code interpretation workflow. Specialists manually read and verify raw grid code files to answer user queries, which is time-consuming and prone to errors.}
\label{fig:conventional}
\end{figure}

Conventional grid code interpretation relies heavily on human expertise, as illustrated in Figure \ref{fig:conventional}. Grid codes are often written with dense, region-specific terminology, technical jargon, and implicit assumptions that demand deep domain knowledge. To navigate these complexities, electricity providers typically consult local specialists or external agencies, which significantly increases operational costs and project timelines. Human interpretation, however, is prone to inconsistency, stemming from ambiguities in the documents or simple human error. The challenge intensifies with lengthy, multi-layered regulations, where manual review becomes prohibitively time-consuming and costly. This dilemma is further compounded for companies operating across multiple jurisdictions, where the limited capacity of specialists creates a critical bottleneck. These challenges underscore the urgent need for scalable, automated solutions that can efficiently process large volumes of regulatory text, deliver reliable compliance interpretation, and reduce the risk of errors.

\begin{table*}[htbp!]
    \centering
    \begin{tabular}{l|c|c|c|c}
    
    \textbf{Method} & \textbf{Domain Adaptability} & \textbf{Deployment Cost} & \textbf{Data Requirement} & \textbf{Long Doc Handling} \\
    \hline
    General LLM       & \ding{55} Low      & \ding{51} Low       & \ding{55} None (pretrained only)  & \ding{55} Poor \\
    Fine-Tuning       & \ding{51} High     & \ding{55} High      & \ding{55} Requires labeled data   & \ding{55} Moderate \\
    RAG               & \ding{51} Medium   & \ding{51} Medium    & \ding{51} Unlabeled documents     & \ding{55} Varies \\
    Improved Retrieval & \ding{51} High    & \ding{51} Medium    & \ding{51} Unlabeled documents     & \ding{51} Strong \\
\end{tabular}
\caption{Comparison of mainstream approaches for industry-specific LLM adaptations.}
\label{tab:adaptation}
\end{table*}

The rapid advancement of large language models (LLMs) \cite{openaiGPT4TechnicalReport2023,IntroducingChatGPT2024,minaeeLargeLanguageModels2025} offers an intriguing solution for grid code reasoning and compliance. With extraordinary text comprehension ability, LLMs excel at text understanding and question answering tasks. However, lack of specialized domain knowledge and susceptibility to hallucinations \cite{xuHallucinationInevitableInnate2025,huangSurveyHallucinationLarge2025} challenge their enterprise and industrial applications. Table \ref{tab:adaptation} have summarized mainstream LLM domain adaptation methods. While finetuning with domain-specific datasets can significantly mitigate some of these issues \cite{jFineTuningLLM2024,jeongFinetuningUtilizationMethods2024}, it requires high-quality datasets, which are difficult to obtain in energy industry due to the data confidentiality and cumbersome work of data washing and formatting. Additionally, substantial computation resource and LLM engineering experience required can hinder the finetuning of LLMs \cite{huLoRALowRankAdaptation2021,chenLongLoRAEfficientFinetuning2024}, especially in mid-size or small enterprises and conventional industries, such as energy and power grids.

Recently, retrieval-augmented generation (RAG) has emerged as a cost-effective alternative to fine-tuning, offering higher accuracy and more flexible deployment for enterprise applications \cite{lewisRetrievalAugmentedGenerationKnowledgeIntensive2021,fanSurveyRAGMeeting2024}. By enabling large language models (LLMs) to seamlessly access external knowledge bases during inference, RAG reduces hallucinations and improves factual consistency without retraining. Constructing such knowledge bases is straightforward: raw documents are split into chunks, embedded, and stored in vector databases—an approach that can be fully deployed locally without extensive data cleaning. As a result, RAG provides a private, scalable, and low-overhead solution for adapting LLMs to industrial use cases \cite{balaguerRAGVsFinetuning2024}.

In this paper, we demonstrate a RAG-driven AI framework for grid code interpretation. We build external knowledge bases with grid code documents and orchestrate them with high performance open-source LLMs, such as DeepSeek \cite{guo2025deepseek, deepseek-aiDeepSeekV3TechnicalReport2025} and Qwen3 \cite{yangQwen3TechnicalReport2025}, which demonstrate strong reasoning ability and scalability. Our system achieves answer quality up to 88\%, as validated by experts from Southern Grid and other external authorities. We believe our framework also holds potential for broader applications beyond compliance interpretation—such as proactively identifying potential regulatory violations and generating power grid simulation configurations. These capabilities could significantly accelerate regulatory workflows and reduce compliance risk in electricity industry.

In this paper, we introduce a RAG-driven AI framework for grid code interpretation, leveraging domain-specific knowledge bases constructed from regulatory documents and orchestrated with high-performance open-source LLMs such as DeepSeek \cite{guo2025deepseek, deepseek-aiDeepSeekV3TechnicalReport2025} and Qwen3 \cite{yangQwen3TechnicalReport2025}. These models exhibit strong reasoning capability and scalability, enabling reliable compliance interpretation. Expert validation with Southern Grid and external authorities shows that our system achieves answer quality of up to 88\%. Beyond compliance interpretation, our framework also holds promise for proactive regulatory monitoring, such as detecting potential violations and generating simulation-ready grid configurations, thereby accelerating regulatory workflows and mitigating compliance risks in the electricity sector.

Our contributions are as follows:
\begin{itemize}
	\item \textbf{First RAG-driven framework for power grid code reasoning and compliance.}
	We present the first RAG-based agent tailored for automated reasoning and compliance verification in grid codes, addressing domain-specific challenges such as technical jargon, implicit assumptions, and regulatory ambiguity.
    \item \textbf{Optimized multi-stage query refinement and enhanced retrieval.}  
    We design a multi-stage refinement pipeline—incorporating term explanation and context injection—that improves query understanding and retrieval precision. This is further integrated with the RAPTOR framework \cite{sarthiRAPTORRECURSIVEABSTRACTIVE2024} to support robust, multi-hop retrieval across lengthy regulatory documents.  
    \item \textbf{Demonstrated accuracy and real-world applicability.}  
    Extensive evaluation shows substantial improvements over baseline LLM and RAG methods in both retrieval accuracy and compliance reasoning, enabling trustworthy, explainable outputs for real-world deployments.

\end{itemize}

\section{Related Work}

\begin{figure*}[htbp!]
\centering
\includegraphics[width=1.8\columnwidth, trim=2cm 1.5cm 2cm 2.3cm, clip]{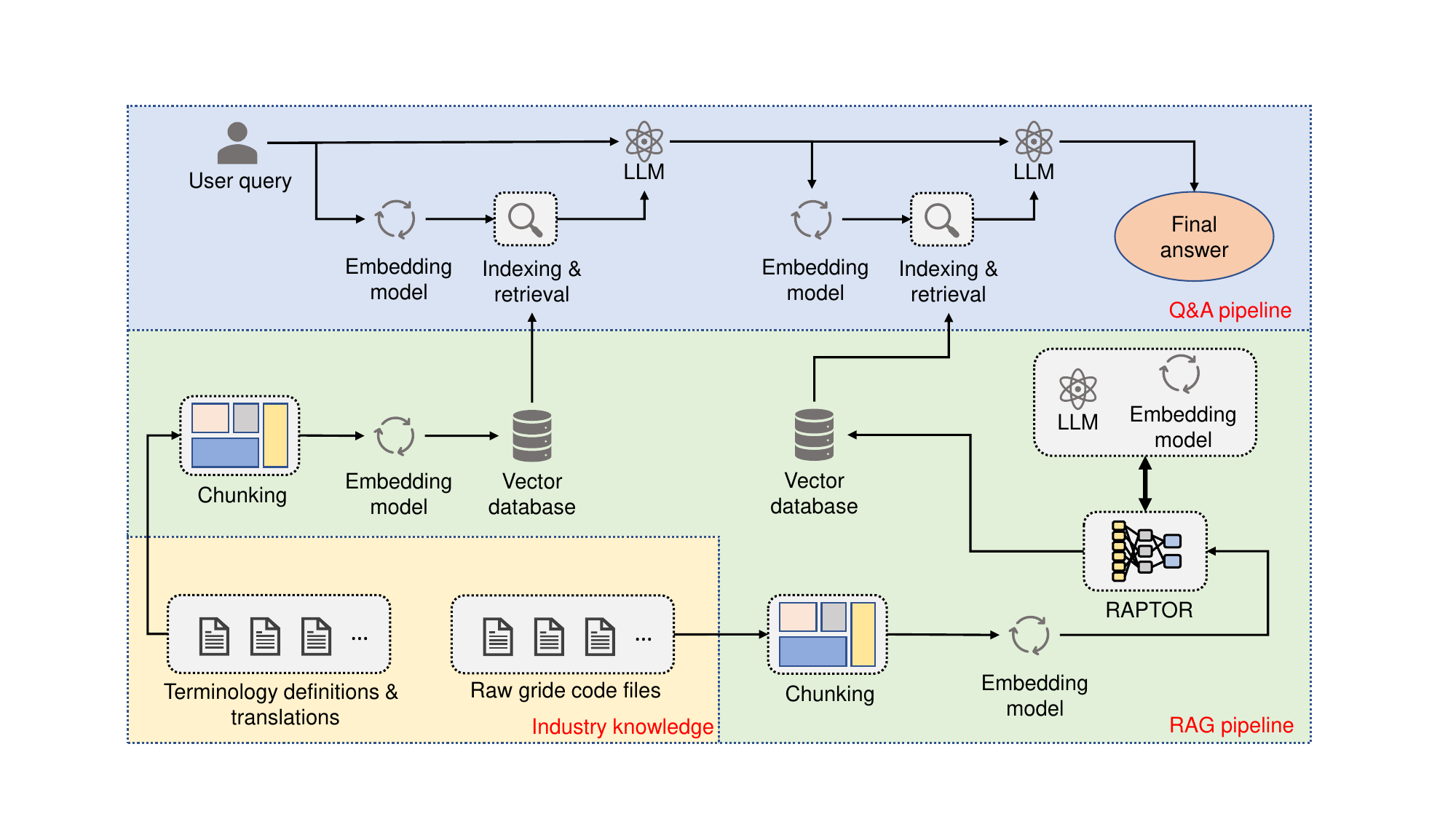} % Reduce the figure size so that it is slightly narrower than the column. Don't use precise values for figure width.This setup will avoid overfull boxes.
\caption{System architecture of GridCodex: The workflow consists of three major components—industry knowledge, a retrieval-augmented generation (RAG) pipeline, and a Q\&A pipeline. Domain-specific terminologies and factual knowledge are processed, embedded, and indexed to construct industry knowledge bases. These knowledge bases support multi-stage query refinement and enable robust compliance interpretation in the Q\&A pipeline.}
\label{fig:arch}
\end{figure*}

\subsection{Large Language Models}

Large Language Models (LLMs) such as ChatGPT \cite{IntroducingChatGPT2024}, Gemini \cite{team2023gemini}, and Qwen \cite{yangQwen3TechnicalReport2025} have demonstrated remarkable success across a wide range of natural language processing (NLP) tasks, including question answering, text summarization, and reasoning. Built on the transformer architecture, LLMs are typically pre-trained on web-scale text corpora with billions of parameters, enabling strong language understanding, generation, and transferability to diverse downstream tasks. More recently, “think-first” models with enhanced reasoning capabilities \cite{guo2025deepseek,yangQwen3TechnicalReport2025} have shown promise in handling complex user requests, decision-making, and external tool invocation, opening new possibilities for fully automated workflows in industrial applications.

Despite these advances, general-purpose LLMs remain vulnerable to hallucinations \cite{xuHallucinationInevitableInnate2025} and often lack the specialized domain knowledge required for enterprise deployment. To address these limitations, various strategies have been explored, including fine-tuning on domain-specific data and retrieval-augmented generation (RAG), which enhances factual reliability and contextual grounding for industry-specific use cases.

\subsection{Retrieval-Augmented Generation (RAG)}

Recently, Retrieval-Augmented Generation (RAG) \cite{lewisRetrievalAugmentedGenerationKnowledgeIntensive2021} has emerged as a practical alternative to fine-tuning for adapting LLMs to domain-specific tasks. By retrieving relevant context from external knowledge bases during inference, RAG enhances LLM performance while avoiding costly and time-intensive model retraining. Advances such as RAPTOR \cite{sarthiRAPTORRECURSIVEABSTRACTIVE2024} and HyDE \cite{gaoPreciseZeroShotDense2022} have further improved retrieval quality for complex document understanding, achieving notable gains in long-context domains such as legal and scientific texts. With its ability to improve factual consistency, reduce hallucinations, and offer greater controllability, RAG is particularly well-suited for knowledge-intensive applications—including the interpretation and compliance verification of power grid codes.

\subsection{LLM applications in the Energy Sector}

Despite the growing interest in LLMs and RAG, their adoption in the power and energy sector remains at an early stage. Recent studies have explored their use in power system question answering—for instance, in dispatch control \cite{zhangIntelligentQASystem2024} and general Q\&A tasks \cite{niChatGridIntelligentKnowledge2024,luRetrievalAugmentedGenerationFramework2024}. However, far less attention has been given to applying LLMs for regulatory interpretation and grid code compliance, a domain where accuracy and reliability are paramount. Existing approaches predominantly rely on supervised fine-tuning to adapt general-purpose LLMs, with limited exploration of model selection and retrieval optimization strategies. To the best of our knowledge, no prior work has systematically employed a RAG-based framework for power grid code reasoning—highlighting a critical gap in both research and practice.

\section{System Architecture}

To enable robust grid code reasoning and compliance verification, GridCodex integrates a tailored workflow that leverages RAG to enhance the accuracy of general-purpose LLMs when responding to grid code queries. As illustrated in Figure \ref{fig:arch}, the system is organized into three core components: industry knowledge, the RAG pipeline, and the Q\&A pipeline. The \textit{industry knowledge} module provides domain-specific terminology, definitions, and translations, enabling LLMs to better interpret user queries. The \textit{RAG pipeline} constructs domain knowledge bases and performs retrieval from external regulatory documents. Finally, the \textit{Q\&A pipeline} interfaces directly with users, orchestrating retrieved context and LLM reasoning to produce reliable and contextually grounded answers.

\subsection{Industry Knowledge}

Industry knowledge (or domain knowledge) is essential for the effective deployment of LLMs in industrial applications. A key limitation of general-purpose LLMs is their lack of specialized expertise, which prevents them from consistently generating accurate answers to domain-specific queries. In the context of grid code compliance, domain knowledge can be broadly classified into two categories: (1) terminology knowledge, including definitions and translations of technical terms, and (2) factual knowledge, consisting of regulatory clauses and provisions from different countries. In GridCodex, these two knowledge types are maintained as separate knowledge bases and directly supplied to the RAG pipeline, eliminating the need for additional preprocessing while ensuring clarity and modularity.

\begin{figure*}[htbp!]
\centering
\includegraphics[width=1.8\columnwidth, trim=1cm 7cm 4cm 3.6cm, clip]{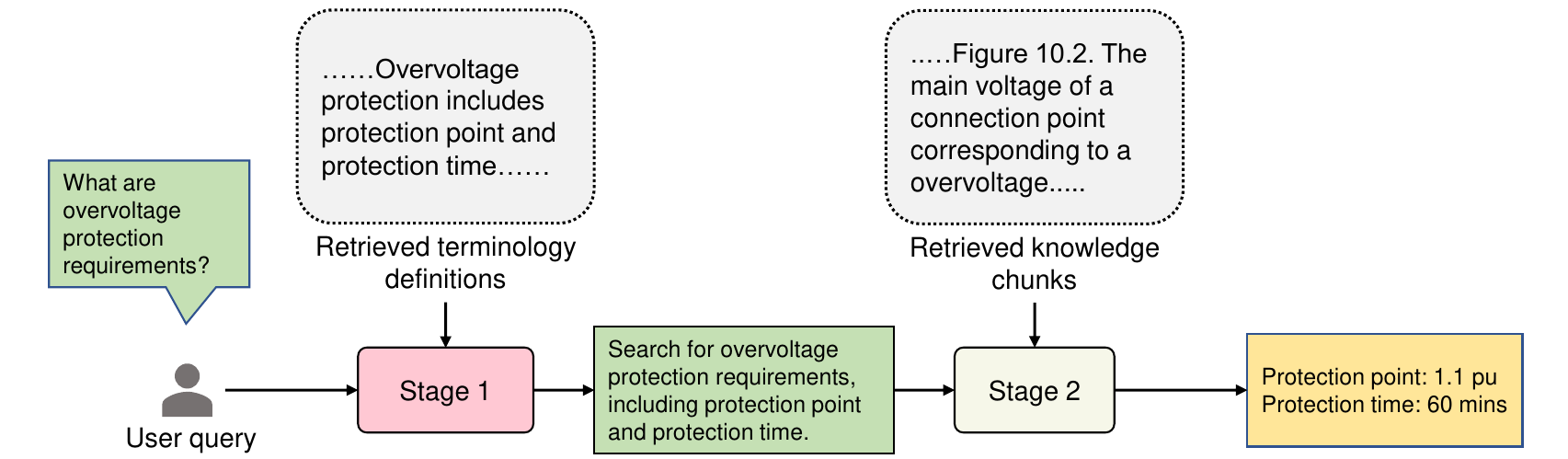} % Reduce the figure size so that it is slightly narrower than the column. Don't use precise values for figure width.This setup will avoid overfull boxes.
\caption{A real use case of GridCodex: In Stage 1, the user query is refined with terminology definitions, then used for raw knowledge retrieval to obtain the final answer.}
\label{fig:example}
\end{figure*}

\subsection{RAG pipeline}

We construct our RAG pipeline using RAGFlow, an open-source framework chosen for its scalability and flexibility. To maximize retrieval accuracy, we process terminology knowledge and factual knowledge differently.

For terminology definitions and translations, we preserve their hierarchical relationships by storing them in JSON and Markdown formats. These files are chunked to match the embedding model’s maximum context window, embedded into dense vectors, and indexed in a vector database.

For factual data—including tables, figures, and raw grid code documents—we employ OCR (Optical Character Recognition), TSR (Table Structure Recognition), and DLR (Document Layout Recognition) as preprocessing steps. Grid codes often contain deeply nested hierarchical structures (e.g., sub-clauses), which pose challenges for retrieval. To preserve semantic integrity, we adopt an adaptive chunking strategy, using the lowest-level section titles as atomic units and preventing clause cutoffs during retrieval. To further capture semantic connections and cross-references, we integrate RAPTOR \cite{sarthiRAPTORRECURSIVEABSTRACTIVE2024}. RAPTOR leverages Gaussian Mixture Models to cluster semantically related chunks, which are then summarized by general-purpose LLMs. These summaries are re-embedded recursively, producing a tree-structured knowledge base that preserves both local detail and global context. The final knowledge bases, covering both terminology and factual data, are stored in vector databases for efficient retrieval.

\subsection{Q\&A Pipeline}

Our Q\&A pipeline follows a classic RAG workflow, in which retrieved knowledge is appended to user queries as contextual input. To improve retrieval accuracy, we employ a multi-stage query refinement strategy. First, terminology knowledge is retrieved and used to enrich the original query with detailed explanations and relevant domain-specific keywords. The refined query is then translated into English—aligning with the predominant language of grid code documents—before being applied to factual knowledge retrieval. This staged rewriting process leverages key terminologies, bridges cross-lingual gaps, and ensures more accurate and robust retrieval. The resulting knowledge, together with the refined English query, is then passed to general-purpose LLMs for answer generation. To further enhance reliability, we incorporate rigorous prompt engineering to guide model reasoning toward precise and truthful responses. Importantly, the Q\&A pipeline is model-agnostic: it can operate with both API-based LLMs and self-hosted open-source models, offering flexibility for different deployment environments.

\begin{table*}[htbp!]
\centering
\caption{Benchmark results for \textit{GridCodex} and baselines.}
\label{tab:benchmark_results}
\begin{tabular}{l|l|c|c|c}
\hline
\textbf{Region} & \textbf{Model} & \textbf{Answer Quality} & \textbf{Faithfulness} & \textbf{Recall@30} \\ \hline
Hong Kong, China & General LLMs & 0.798 & -- & -- \\
               & Vanilla RAG & 0.759 & 0.978 & 0.182 \\
               & Optimized RAG (\textit{GridCodex}) & \textbf{0.946} & \textbf{0.978} & \textbf{1.000} \\ \hline
Netherlands & General LLMs & 0.602 & -- & -- \\
                 & Vanilla RAG & 0.519 & \textbf{0.978} & 0.100 \\
                 & Optimized RAG (\textit{GridCodex}) & \textbf{0.852} & 0.968 & \textbf{0.917} \\ \hline
European Union & General LLMs & 0.602 & -- & -- \\
                    & Vanilla RAG & 0.645 & 0.941 & 0.100 \\
                    & Optimized RAG (\textit{GridCodex}) & \textbf{0.843} & \textbf{0.968} & \textbf{0.913} \\ \hline
Bangladesh  & General LLMs & 0.784 & -- & -- \\
                & Vanilla RAG & 0.805 & \textbf{0.984} & 0.000 \\
                & Optimized RAG (\textit{GridCodex}) & \textbf{0.877} & 0.962 & \textbf{0.900} \\ \hline
\end{tabular}
\end{table*}
\section{Experiments}
\subsection{Experiment Setup}

We evaluate \textit{GridCodex} against two baselines:  
\begin{itemize}
	\item \textbf{Grid code compliance with general LLMs:} Queries are directly submitted to general-purpose LLMs with minimal prompt engineering and region-specific metadata. Guardrails are included to suppress hallucinations when the model encounters queries outside its knowledge.  
	\item \textbf{Grid code compliance with vanilla RAG:} This baseline incorporates RAPTOR-enabled knowledge bases \cite{sarthiRAPTORRECURSIVEABSTRACTIVE2024}. Questions are used to retrieve relevant chunks, which are appended to the user queries before being processed by a general LLM. Prompt-level guardrails are added to mitigate hallucinations when retrieval fails.  
\end{itemize}

We evaluate system performance on three key metrics:  
\begin{itemize}
	\item \textbf{Answer Quality:} It evaluates \textit{accuracy} (correctness), \textit{completeness} (coverage of essential points), and \textit{usefulness} (relevance to the query). Accuracy is prioritized, with completeness and usefulness downweighted when accuracy is low.  
	\item \textbf{Faithfulness:} It measures consistency between the generated answers and the retrieved knowledge chunks.  
	\item \textbf{Recall@30:} It assesses whether the information necessary to answer a query is contained within the top 30 retrieved chunks.  
\end{itemize}

To ensure fairness and objectivity, we adopt automated scoring using LLM-based evaluators. Carefully designed prompts and expert-provided reference answers from the electricity industry guide the evaluation.

\subsection{Datasets}

We benchmark \textit{GridCodex} using a proprietary HUAWEI grid code dataset comprising question–answer pairs derived from official documents issued by four regulatory agencies: Hong Kong (China), the Netherlands, the European Union, and Bangladesh. The dataset contains 148 QA pairs in total, covering diverse scenarios of grid code interpretation and compliance verification.

\subsection{Models}

The \textit{GridCodex} pipeline integrates carefully selected LLMs and embedding models, chosen to balance retrieval accuracy, reasoning ability, answer quality, and computational efficiency:  
\begin{itemize}
	\item \textbf{Embedding model:} \texttt{Linq-Embed-Mistral} \cite{choiLinqEmbedMistralTechnicalReport2024} is used to construct vector embeddings of knowledge chunks. Trained on high-quality synthetic data, it delivers strong precision and reliability in semantic search.  
	\item \textbf{RAPTOR summarization:} \texttt{DeepSeek-R1-0528} \cite{guo2025deepseek} is employed for hierarchical document summarization, leveraging strong reasoning capabilities for semantic clustering.  
	\item \textbf{Query refinement and translation:} For terminology expansion and translation, we adopt \texttt{Qwen3-32B-AWQ} \cite{yangQwen3TechnicalReport2025}, which balances refinement quality and hardware efficiency.  
	\item \textbf{Answer synthesis and scoring:} \texttt{Qwen3-235B-A22B} \cite{yangQwen3TechnicalReport2025}, a mixture-of-experts (MoE) model, is used for final answer generation and automated scoring, providing concise, faithful responses.  
\end{itemize}

\subsection{Results}

Table~\ref{tab:benchmark_results} presents the results across four regulatory regions. Since the plain LLM baseline lacks retrieval, only answer quality is reported. Vanilla RAG provides moderate improvements in certain cases (e.g., the EU and Bangladesh datasets), but its performance is inconsistent, particularly for Hong Kong and the Netherlands. These deficiencies are largely attributable to language mismatches and terminology ambiguities: vanilla RAG frequently retrieves noisy or partially relevant chunks, reflected in its low Recall@30 scores.

In contrast, \textit{GridCodex} consistently outperforms both baselines across all evaluation metrics. Through multi-stage query refinement and RAPTOR-based retrieval, it achieves high and stable document coverage (Recall@30 $>0.90$ in all regions) while generating accurate and faithful answers. For example, in the Hong Kong (China) dataset, \textit{GridCodex} achieves perfect retrieval coverage (Recall@30 = 1.0) and a remarkable answer quality score of 0.946, yielding an overall F1 of 0.972.  

Similar gains are observed for the Netherlands and EU datasets, which are more challenging due to complex, data-heavy documents. Here, \textit{GridCodex} improves answer quality by more than 30\% relative to vanilla RAG, while maintaining strong faithfulness ($\sim$0.968) and high retrieval coverage (Recall@30 = 0.917 and 0.913, respectively). For Bangladesh, where the grid code documents are shorter but contain under-defined clauses, \textit{GridCodex} still yields significant improvements, raising answer quality from 0.805 to 0.877 with Recall@30 maintained at 0.900.

\begin{figure}
	\centering
	\includegraphics[width=0.8\columnwidth, trim=0cm 0cm 0cm 0cm, clip]{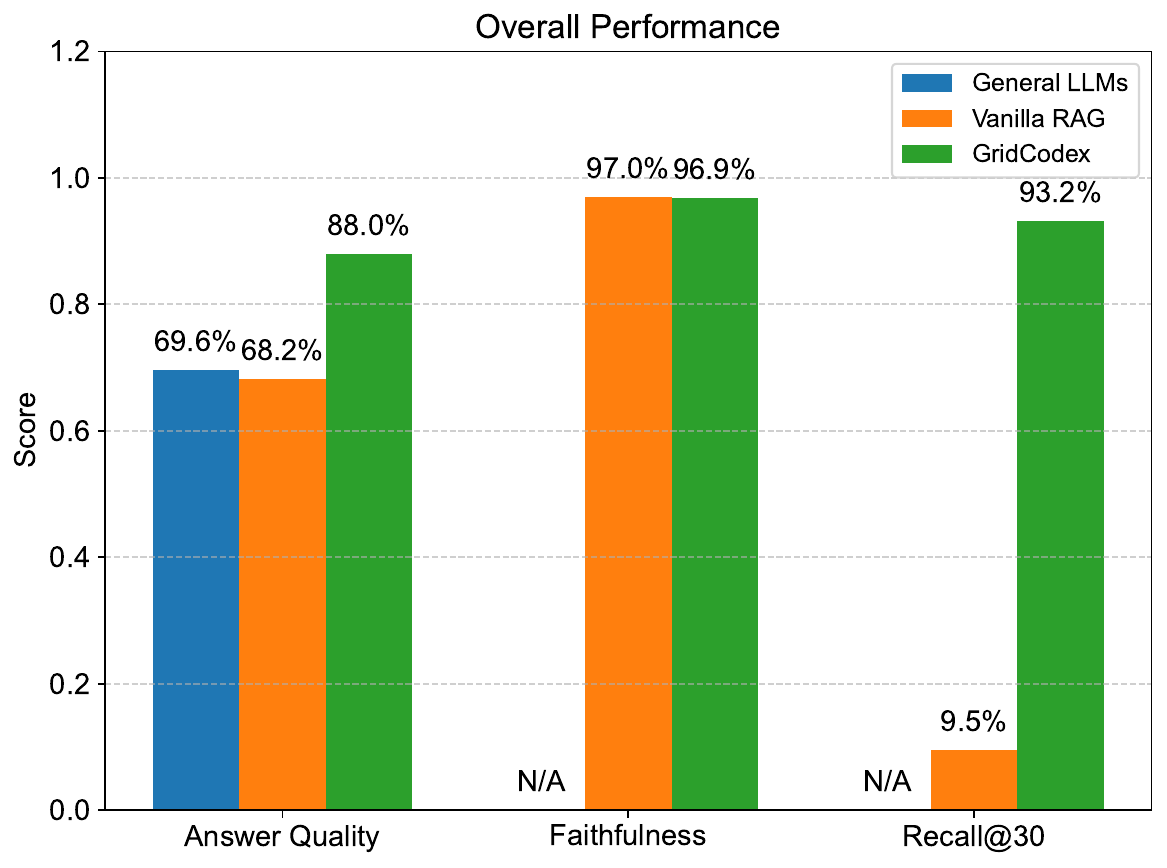} 
	\caption{Overall performance of different systems across regions. (Subject to aesthetic changes.)}
	\label{fig:overall}
\end{figure}

As shown in Figure~\ref{fig:overall}, \textit{GridCodex} delivers an overall 27.5\% improvement in answer quality compared to the two baselines. These gains stem from its multi-stage query refinement and RAPTOR-enhanced multi-hop retrieval. The framework also achieves high faithfulness through careful model selection and prompt engineering. Most notably, \textit{GridCodex} increases Recall@30 by nearly 9$\times$ compared with the vanilla RAG system, establishing itself as a robust end-to-end solution for grid code reasoning and compliance.  

\section{Ablation Study: Model Size and Reasoning Capability}

We conduct an ablation study to assess how model size and reasoning capability influence \textit{GridCodex} performance. 
All experiments are performed on the Netherlands dataset using the default \textit{GridCodex} configuration. To isolate the effect of the LLMs, we keep query refinement, embedding models, and RAPTOR settings fixed across comparisons.

\subsection{Model Size: Large vs.\ Mid-size}
To study the role of scale, we compare:
(1) \texttt{Qwen3-235B-A22B} \cite{yangQwen3TechnicalReport2025}, the flagship model with 235 billion total parameters and 22 billion active, optimized for long-context reasoning, and  
(2) \texttt{Qwen3-30B-A3B} \cite{yangQwen3TechnicalReport2025}, a 30B-parameter variant with 3B active, designed for faster inference and reduced hardware costs.

\begin{table}[h]
	\centering
	\caption{Model size impact (Netherlands dataset).}
	\label{tab:size_impact}
	\begin{tabular}{l|c|c}
		\hline
		\textbf{Model} & \textbf{Answer Quality} & \textbf{Faithfulness} \\ \hline
		235B-A22B & \textbf{0.852} & 0.968 \\ 
		30B-A3B   & 0.740 & \textbf{0.995} \\ 
		\hline
	\end{tabular}
\end{table}

As shown in Table~\ref{tab:size_impact}, scaling from 30B to 235B parameters improves answer quality by 11.2\%. The larger model demonstrates stronger multi-hop reasoning, cross-clause referencing, and ambiguity resolution due to greater representational capacity and expert routing.  
However, this gain comes at a cost: faithfulness drops slightly (0.968 vs.\ 0.995), as the 235B model integrates retrieved context more aggressively, occasionally introducing inferred connections not explicitly present in the documents. This highlights a key trade-off—larger models enable richer reasoning but may lean toward speculative integration, whereas smaller models remain more conservative and strictly grounded in retrieved evidence.

\subsection{Reasoning Capability: Thinking vs.\ Non-Thinking}
Next, we examine the effect of reasoning ability by comparing two variants of the same 235B model:  
(1) \texttt{Qwen3-235B-A22B} \cite{yangQwen3TechnicalReport2025}, the default reasoning-capable flagship, and  
(2) \texttt{Qwen3-235B-A22B-Instruct-2507} \cite{yangQwen3TechnicalReport2025}, an instruction-following variant optimized for direct answer generation with reasoning chains disabled.

\begin{table}[h]
	\centering
	\caption{Impact of reasoning optimization (Netherlands dataset).}
	\label{tab:thinking_vs_instruct}
	\begin{tabular}{lcc}
		\hline
		\textbf{Model} & \textbf{Answer Quality} & \textbf{Faithfulness} \\ \hline
		235B & \textbf{0.852} & 0.968 \\ 
		235B-Instruct & 0.795 & \textbf{0.989} \\ 
		\hline
	\end{tabular}
\end{table}

Table~\ref{tab:thinking_vs_instruct} shows that the reasoning-enabled model delivers a 5.8\% improvement in answer quality. Reasoning capability proves essential for deep contextual linking, enabling the model to resolve implicit dependencies and navigate dense regulatory cross-references. Conversely, the Instruct variant, while slightly weaker in reasoning, achieves higher faithfulness by adhering more strictly to retrieved evidence.

\subsection{Key Observations}
The ablation highlights two major findings:
\begin{enumerate}
	\item \textbf{Model scale} boosts reasoning and ambiguity resolution but can introduce speculative interpretations that reduce strict faithfulness.  
	\item \textbf{Reasoning ability} enhances clause linking and implicit dependency resolution, albeit with a tendency toward more interpretive answers.  
\end{enumerate}
For real-world deployments, mid-sized reasoning models (e.g., \texttt{Qwen3-30B-A3B}) represent a strong balance—retaining much of the reasoning power of very large models while being computationally efficient.

\section{Lessons Learned}
From developing and evaluating \textit{GridCodex}, we distill several practical insights:

\begin{itemize}
	\item \textbf{Better queries lead to better answers.}  
	Grid codes contain ambiguous references, nested clauses, and dense jargon, making both retrieval and QA challenging. Direct retrieval on raw user queries often fails to capture the most relevant knowledge chunks, yielding incomplete context. Moreover, without explicit clarification of key terms, LLMs struggle to interpret queries or fully leverage retrieved knowledge. Our multi-stage query refinement—combining terminology explanation, rewriting, and translation—significantly improves retrieval coverage and overall answer quality.  
	
	\item \textbf{Regulatory QA requires reasoning beyond surface text.}  
	Regulatory documents are not simple fact repositories; they demand reasoning across implicit dependencies and extensive cross-references. Our experiments show that reasoning-oriented models consistently outperform purely instruction-following models, underlining the necessity of reasoning ability for compliance interpretation.  
\end{itemize}

These lessons suggest that effective grid code reasoning systems require careful integration of algorithmic design (retrieval and reasoning) with engineering optimizations (query refinement and model selection).

\section{Conclusion}

This paper presented \textit{GridCodex}, a retrieval-augmented generation (RAG) framework designed for reasoning and compliance verification in power grid codes. By combining domain-specialized knowledge bases, multi-stage query refinement, and high-performance open-source LLMs, \textit{GridCodex} delivers significant gains in both answer quality and reliability. 
Extensive experiments demonstrate that \textit{GridCodex} effectively interprets technical terminologies, nested clauses, and complex cross-references, enabling accurate grid code reasoning and compliance assessment. Beyond power grids, the framework shows strong potential for proactive violation detection, automated infrastructure configuration, and broader compliance workflows in other safety-critical domains.

\bibliography{aaai2026}

@article{ullahDynamicPerformancePower2025,
  title = {Dynamic {{Performance}} and {{Power Quality}} of {{Large-Scale Wind Power Plants}}: {{A Review}} on {{Challenges}}, {{Evolving Grid Code}}, and {{Proposed Solutions}}},
  shorttitle = {Dynamic {{Performance}} and {{Power Quality}} of {{Large-Scale Wind Power Plants}}},
  author = {Ullah, Mohib and Guan, Yajuan and Yu, Yun and Chaudhary, Sanjay K. and Vasquez, Juan C. and Guerrero, Josep M.},
  year = {2025},
  journal = {IEEE Open Journal of Power Electronics},
  volume = {6},
  pages = {1148--1173},
  issn = {2644-1314},
  doi = {10.1109/OJPEL.2025.3582012},
  urldate = {2025-08-06}
}

@article{GridCodeReinforcements2016,
  title = {Grid Code Reinforcements for Deeper Renewable Generation in Insular Energy Systems},
  year = {2016},
  month = jan,
  journal = {Renewable and Sustainable Energy Reviews},
  volume = {53},
  pages = {163--177},
  publisher = {Pergamon},
  issn = {1364-0321},
  doi = {10.1016/j.rser.2015.08.047},
  urldate = {2025-08-06},
  langid = {american}
}

@incollection{RenewablesIntegrationIslands2017,
  title = {Renewables {{Integration}} on {{Islands}}},
  booktitle = {Renewable {{Energy Integration}}},
  year = {2017},
  month = jan,
  pages = {319--329},
  publisher = {Academic Press},
  doi = {10.1016/B978-0-12-809592-8.00024-X},
  urldate = {2025-08-06},
  langid = {american}
}

@inproceedings{ntomarisReserveQuantificationInsular2014,
  title = {Reserve Quantification in Insular Power Systems with High Wind Penetration},
  booktitle = {{{IEEE PES Innovative Smart Grid Technologies}}, {{Europe}}},
  author = {Ntomaris, Andreas V. and Bakirtzis, Emmanouil A. and Chatzigiannis, Dimitris I. and Simoglou, Christos K. and Biskas, Pandelis N. and Bakirtzis, Anastasios G.},
  year = {2014},
  month = oct,
  pages = {1--6},
  issn = {2165-4824},
  doi = {10.1109/ISGTEurope.2014.7028907},
  urldate = {2025-08-06}
}

@misc{IntroducingChatGPT2024,
  title = {Introducing {{ChatGPT}}},
  year = {2024},
  month = mar,
  urldate = {2025-08-06},
  howpublished = {https://openai.com/index/chatgpt/},
  langid = {american}
}

@misc{minaeeLargeLanguageModels2025,
  title = {Large {{Language Models}}: {{A Survey}}},
  shorttitle = {Large {{Language Models}}},
  author = {Minaee, Shervin and Mikolov, Tomas and Nikzad, Narjes and Chenaghlu, Meysam and Socher, Richard and Amatriain, Xavier and Gao, Jianfeng},
  year = {2025},
  month = mar,
  number = {arXiv:2402.06196},
  eprint = {2402.06196},
  primaryclass = {cs},
  publisher = {arXiv},
  doi = {10.48550/arXiv.2402.06196},
  urldate = {2025-08-06},
  archiveprefix = {arXiv}
}

@misc{openaiGPT4TechnicalReport2023,
  title = {{{GPT-4 Technical Report}}},
  author = {OpenAI},
  year = {2023},
  month = mar,
  number = {arXiv:2303.08774},
  eprint = {2303.08774},
  primaryclass = {cs},
  publisher = {arXiv},
  doi = {10.48550/arXiv.2303.08774},
  urldate = {2025-08-06},
  archiveprefix = {arXiv}
}

@article{huangSurveyHallucinationLarge2025,
  title = {A {{Survey}} on {{Hallucination}} in {{Large Language Models}}: {{Principles}}, {{Taxonomy}}, {{Challenges}}, and {{Open Questions}}},
  shorttitle = {A {{Survey}} on {{Hallucination}} in {{Large Language Models}}},
  author = {Huang, Lei and Yu, Weijiang and Ma, Weitao and Zhong, Weihong and Feng, Zhangyin and Wang, Haotian and Chen, Qianglong and Peng, Weihua and Feng, Xiaocheng and Qin, Bing and Liu, Ting},
  year = {2025},
  month = mar,
  journal = {ACM Transactions on Information Systems},
  volume = {43},
  number = {2},
  eprint = {2311.05232},
  primaryclass = {cs},
  pages = {1--55},
  issn = {1046-8188, 1558-2868},
  doi = {10.1145/3703155},
  urldate = {2025-08-06},
  archiveprefix = {arXiv}
}

@misc{xuHallucinationInevitableInnate2025,
  title = {Hallucination Is {{Inevitable}}: {{An Innate Limitation}} of {{Large Language Models}}},
  shorttitle = {Hallucination Is {{Inevitable}}},
  author = {Xu, Ziwei and Jain, Sanjay and Kankanhalli, Mohan},
  year = {2025},
  month = feb,
  number = {arXiv:2401.11817},
  eprint = {2401.11817},
  primaryclass = {cs},
  publisher = {arXiv},
  doi = {10.48550/arXiv.2401.11817},
  urldate = {2025-08-06},
  archiveprefix = {arXiv}
}

@article{jeongFinetuningUtilizationMethods2024,
  title = {Fine-Tuning and {{Utilization Methods}} of {{Domain-specific LLMs}}},
  author = {Jeong, Cheonsu},
  year = {2024},
  month = mar,
  journal = {Journal of Intelligence and Information Systems},
  volume = {30},
  number = {1},
  eprint = {2401.02981},
  primaryclass = {cs},
  pages = {93--120},
  issn = {2288-4866, 2288-4882},
  doi = {10.13088/jiis.2024.30.1.093},
  urldate = {2025-08-07},
  archiveprefix = {arXiv}
}

@misc{jFineTuningLLM2024,
  title = {Fine {{Tuning LLM}} for {{Enterprise}}: {{Practical Guidelines}} and {{Recommendations}}},
  shorttitle = {Fine {{Tuning LLM}} for {{Enterprise}}},
  author = {J, Mathav Raj and VM, Kushala and Warrier, Harikrishna and Gupta, Yogesh},
  year = {2024},
  month = mar,
  number = {arXiv:2404.10779},
  eprint = {2404.10779},
  primaryclass = {cs},
  publisher = {arXiv},
  doi = {10.48550/arXiv.2404.10779},
  urldate = {2025-08-07},
  archiveprefix = {arXiv}
}

@misc{huLoRALowRankAdaptation2021,
  title = {{{LoRA}}: {{Low-Rank Adaptation}} of {{Large Language Models}}},
  shorttitle = {{{LoRA}}},
  author = {Hu, Edward J. and Shen, Yelong and Wallis, Phillip and {Allen-Zhu}, Zeyuan and Li, Yuanzhi and Wang, Shean and Wang, Lu and Chen, Weizhu},
  year = {2021},
  month = oct,
  number = {arXiv:2106.09685},
  eprint = {2106.09685},
  primaryclass = {cs},
  publisher = {arXiv},
  doi = {10.48550/arXiv.2106.09685},
  urldate = {2025-08-07},
  archiveprefix = {arXiv}
}

@misc{chenLongLoRAEfficientFinetuning2024,
  title = {{{LongLoRA}}: {{Efficient Fine-tuning}} of {{Long-Context Large Language Models}}},
  shorttitle = {{{LongLoRA}}},
  author = {Chen, Yukang and Qian, Shengju and Tang, Haotian and Lai, Xin and Liu, Zhijian and Han, Song and Jia, Jiaya},
  year = {2024},
  month = mar,
  number = {arXiv:2309.12307},
  eprint = {2309.12307},
  primaryclass = {cs},
  publisher = {arXiv},
  doi = {10.48550/arXiv.2309.12307},
  urldate = {2025-08-07},
  archiveprefix = {arXiv}
}

@misc{fanSurveyRAGMeeting2024,
  title = {A {{Survey}} on {{RAG Meeting LLMs}}: {{Towards Retrieval-Augmented Large Language Models}}},
  shorttitle = {A {{Survey}} on {{RAG Meeting LLMs}}},
  author = {Fan, Wenqi and Ding, Yujuan and Ning, Liangbo and Wang, Shijie and Li, Hengyun and Yin, Dawei and Chua, Tat-Seng and Li, Qing},
  year = {2024},
  month = jun,
  number = {arXiv:2405.06211},
  eprint = {2405.06211},
  primaryclass = {cs},
  publisher = {arXiv},
  doi = {10.48550/arXiv.2405.06211},
  urldate = {2025-08-07},
  archiveprefix = {arXiv}
}

@misc{lewisRetrievalAugmentedGenerationKnowledgeIntensive2021,
  title = {Retrieval-{{Augmented Generation}} for {{Knowledge-Intensive NLP Tasks}}},
  author = {Lewis, Patrick and Perez, Ethan and Piktus, Aleksandra and Petroni, Fabio and Karpukhin, Vladimir and Goyal, Naman and K{\"u}ttler, Heinrich and Lewis, Mike and Yih, Wen-tau and Rockt{\"a}schel, Tim and Riedel, Sebastian and Kiela, Douwe},
  year = {2021},
  month = apr,
  number = {arXiv:2005.11401},
  eprint = {2005.11401},
  primaryclass = {cs},
  publisher = {arXiv},
  doi = {10.48550/arXiv.2005.11401},
  urldate = {2025-08-07},
  archiveprefix = {arXiv}
}

@misc{balaguerRAGVsFinetuning2024,
  title = {{{RAG}} vs {{Fine-tuning}}: {{Pipelines}}, {{Tradeoffs}}, and a {{Case Study}} on {{Agriculture}}},
  shorttitle = {{{RAG}} vs {{Fine-tuning}}},
  author = {Balaguer, Angels and Benara, Vinamra and Cunha, Renato Luiz de Freitas and Filho, Roberto de M. Estev{\~a}o and Hendry, Todd and Holstein, Daniel and Marsman, Jennifer and Mecklenburg, Nick and Malvar, Sara and Nunes, Leonardo O. and Padilha, Rafael and Sharp, Morris and Silva, Bruno and Sharma, Swati and Aski, Vijay and Chandra, Ranveer},
  year = {2024},
  month = jan,
  number = {arXiv:2401.08406},
  eprint = {2401.08406},
  primaryclass = {cs},
  publisher = {arXiv},
  doi = {10.48550/arXiv.2401.08406},
  urldate = {2025-08-07},
  archiveprefix = {arXiv}
}

@article{guo2025deepseek,
  title={Deepseek-r1: Incentivizing reasoning capability in llms via reinforcement learning},
  author={Guo, Daya and Yang, Dejian and Zhang, Haowei and Song, Junxiao and Zhang, Ruoyu and Xu, Runxin and Zhu, Qihao and Ma, Shirong and Wang, Peiyi and Bi, Xiao and others},
  journal={arXiv preprint arXiv:2501.12948},
  year={2025}
}

@misc{deepseek-aiDeepSeekV3TechnicalReport2025,
  title={Deepseek-v3 technical report},
  author={Liu, Aixin and Feng, Bei and Xue, Bing and Wang, Bingxuan and Wu, Bochao and Lu, Chengda and Zhao, Chenggang and Deng, Chengqi and Zhang, Chenyu and Ruan, Chong and others},
  journal={arXiv preprint arXiv:2412.19437},
  year={2024}
}

@misc{yangQwen3TechnicalReport2025,
  title={Qwen3 technical report},
  author={Yang, An and Li, Anfeng and Yang, Baosong and Zhang, Beichen and Hui, Binyuan and Zheng, Bo and Yu, Bowen and Gao, Chang and Huang, Chengen and Lv, Chenxu and others},
  journal={arXiv preprint arXiv:2505.09388},
  year={2025}
}

@article{sarthiRAPTORRECURSIVEABSTRACTIVE2024,
  title = {{{RAPTOR}}: {{RECURSIVE ABSTRACTIVE PROCESSING FOR TREE-ORGANIZED RETRIEVAL}}},
  author = {Sarthi, Parth and Abdullah, Salman and Tuli, Aditi and Khanna, Shubh and Goldie, Anna and Manning, Christopher D},
  year = {2024},
  langid = {english}
}

@article{team2023gemini,
  title={Gemini: a family of highly capable multimodal models},
  author={Team, Gemini and Anil, Rohan and Borgeaud, Sebastian and Alayrac, Jean-Baptiste and Yu, Jiahui and Soricut, Radu and Schalkwyk, Johan and Dai, Andrew M and Hauth, Anja and Millican, Katie and others},
  journal={arXiv preprint arXiv:2312.11805},
  year={2023}
}

@inproceedings{luRetrievalAugmentedGenerationFramework2024,
  title = {A {{Retrieval-Augmented Generation Framework}} for {{Electric Power Industry Question Answering}}},
  booktitle = {Proceedings of the 2024 2nd {{International Conference}} on {{Electronics}}, {{Computers}} and {{Communication Technology}}},
  author = {Lu, Yanyan and Peng, Jiao and Xu, Xing and He, Yue and Li, Tao and Wei, Jie and Jing, Hongyu and Wang, Heqing and Xu, Bo and Song, Hui},
  year = {2024},
  month = oct,
  pages = {95--100},
  publisher = {ACM},
  address = {Chengdu China},
  doi = {10.1145/3705754.3705771},
  urldate = {2025-08-07},
  isbn = {979-8-4007-1019-3},
  langid = {english}
}

@inproceedings{niChatGridIntelligentKnowledge2024,
  title = {{{ChatGrid}}: {{Intelligent Knowledge Q}}\&{{A}} for {{Power Dispatching Control Based}} on {{Large Language Models}} and {{Retrieval-augmented Generation}}},
  shorttitle = {{{ChatGrid}}},
  booktitle = {2024 {{IEEE}} 7th {{Information Technology}}, {{Networking}}, {{Electronic}} and {{Automation Control Conference}} ({{ITNEC}})},
  author = {Ni, Mingjian and Zhang, Jing and Fu, Cong and Wang, Jiaqi and Ning, Xin and Li, Siyuan},
  year = {2024},
  month = sep,
  volume = {7},
  pages = {921--925},
  issn = {2693-3128},
  doi = {10.1109/ITNEC60942.2024.10733337},
  urldate = {2025-08-07}
}

@inproceedings{zhangIntelligentQASystem2024,
  title = {Intelligent {{Q}}\&{{A System}} for {{Power Dispatching Based}} on {{Retrieval Augmented Generation}}},
  booktitle = {2024 6th {{International Conference}} on {{Energy}}, {{Power}} and {{Grid}} ({{ICEPG}})},
  author = {Zhang, Kai and Li, Lei and Xu, Xi and Xin, Rui and Xu, Xing and Zhang, Pengfei},
  year = {2024},
  month = sep,
  pages = {1604--1608},
  doi = {10.1109/ICEPG63230.2024.10775438},
  urldate = {2025-08-07}
}

@misc{gaoPreciseZeroShotDense2022,
  title = {Precise {{Zero-Shot Dense Retrieval}} without {{Relevance Labels}}},
  author = {Gao, Luyu and Ma, Xueguang and Lin, Jimmy and Callan, Jamie},
  year = {2022},
  month = dec,
  number = {arXiv:2212.10496},
  eprint = {2212.10496},
  primaryclass = {cs},
  publisher = {arXiv},
  doi = {10.48550/arXiv.2212.10496},
  urldate = {2025-08-07},
  archiveprefix = {arXiv}
}

@misc{choiLinqEmbedMistralTechnicalReport2024,
  title = {Linq-{{Embed-Mistral Technical Report}}},
  author = {Choi, Chanyeol and Kim, Junseong and Lee, Seolhwa and Kwon, Jihoon and Gu, Sangmo and Kim, Yejin and Cho, Minkyung and Sohn, Jy-yong},
  year = {2024},
  month = dec,
  number = {arXiv:2412.03223},
  eprint = {2412.03223},
  primaryclass = {cs},
  publisher = {arXiv},
  doi = {10.48550/arXiv.2412.03223},
  urldate = {2025-08-07},
  archiveprefix = {arXiv}
}

\end{document}